\documentclass[twocolumn]{IEEEtran}
\usepackage{lscape}
\usepackage{booktabs}
\usepackage{authblk}
\usepackage{url}
\usepackage{silence}
\usepackage[pdftex]{graphicx}
\usepackage{epstopdf}
\usepackage{pifont}
\usepackage{multirow}
\usepackage{float}
\usepackage{caption}
\usepackage{subcaption}
\usepackage[table,xcdraw]{xcolor}
\usepackage{array}
\usepackage{subcaption}
\usepackage{verse}
\usepackage{longtable}
\usepackage{supertabular}
\usepackage{cite}
\usepackage{color}
\usepackage{amsmath}
\usepackage{lettrine}
\usepackage{hyperref}
\usepackage{multirow}
\usepackage{multicol}
\usepackage{dirtytalk}
\usepackage{color}

\usepackage[colorinlistoftodos,prependcaption,textsize=tiny]{todonotes}
\usepackage{booktabs}
\usepackage[normalem]{ulem}
\useunder{\uline}{\ul}{}

\hypersetup{
    colorlinks=false,
    linkcolor=blue,
    filecolor=magenta,
    urlcolor=cyan,
}

\begin{document}

\title{Sentiment Analysis of Users' Reviews on COVID-19 Contact Tracing Apps with a Benchmark Dataset} %-- FA i am voting for this

\author{
    \IEEEauthorblockN{Kashif Ahmad\IEEEauthorrefmark{1}, Firoj Alam\IEEEauthorrefmark{2}, Junaid Qadir\IEEEauthorrefmark{3},
    Basheer Qolomany\IEEEauthorrefmark{4},
    Imran Khan\IEEEauthorrefmark{5}, Talhat Khan\IEEEauthorrefmark{5}, Muhammad Suleman\IEEEauthorrefmark{5}, Naina Said\IEEEauthorrefmark{5}, Syed Zohaib Hassan\IEEEauthorrefmark{6}, Asma Gul\IEEEauthorrefmark{7}, Ala Al-Fuqaha\IEEEauthorrefmark{1}
    }
    \\
    \IEEEauthorblockA{
    \IEEEauthorrefmark{1}College of Science and Engineering, HBKU, Doha, Qatar \\
    \IEEEauthorrefmark{2}Qatar Computing Research Institute, HBKU, Qatar \\
    \IEEEauthorrefmark{3}Dept.of  EE, Information Technology University, Lahore, Pakistan \\
    \IEEEauthorrefmark{4}Dept. of Cyber Systems, University of Nebraska,  Kearney, US\\
    \IEEEauthorrefmark{5}DCSE, University of Engineering and Technology, Peshawar, Pakistan \\
    \IEEEauthorrefmark{6}University of Trento, Italy\\
    \IEEEauthorrefmark{7} Dept. of Stat., Shaheed Benazir Bhutto Women University, Peshawar, Pakistan\\
    \IEEEauthorrefmark{1}\{kahmad,aalfuqaha\}@hbku.edu.qa} \IEEEauthorrefmark{2}fialam@hbku.edu.qa, \IEEEauthorrefmark{3}junaid.qadir@itu.edu.pk, \IEEEauthorrefmark{4}qolomanyb@unk.edu, \IEEEauthorrefmark{5}imrancaan@outlook.com, \IEEEauthorrefmark{5}talhatkhan95@gmail.com, \IEEEauthorrefmark{5}nainasaid@uetpeshawar.edu.pk, \IEEEauthorrefmark{6}zohaib.tirmazi@gmail.com, \IEEEauthorrefmark{7}asmagul@sbbwu.edu.pk
}
\maketitle

\begin{abstract}
Contact tracing has been globally adopted in the fight to control the infection rate of COVID-19. Thanks to digital technologies, such as smartphones and wearable devices, contacts of COVID-19 patients can be easily traced and informed about their potential exposure to the virus. To this aim, several interesting mobile applications have been developed. However, there are ever-growing concerns over the working mechanism and performance of these applications. The literature already provides some interesting exploratory studies on the community's response to the applications by analyzing information from different sources, such as news and users' reviews of the applications. However, to the best of our knowledge, there is no existing solution that automatically analyzes users' reviews and extracts the evoked sentiments. In this work, we propose a pipeline starting from manual annotation via a crowd-sourcing study and concluding on the development and training of AI models for automatic sentiment analysis of users' reviews. In total, we employ eight different methods achieving up to an average F1-Scores 94.8\% indicating the feasibility of automatic sentiment analysis of users' reviews on the COVID-19 contact tracing applications.  We also highlight the key advantages, drawbacks, and users' concerns over the applications. Moreover, we also collect and annotate a large-scale %international 
dataset composed of 34,534 reviews manually annotated from the contract tracing applications of 46 distinct countries. The presented analysis and the dataset are expected to provide a baseline/benchmark for future research in the domain. 

\end{abstract}

\section{Introduction}
Since the emerge of COVID-19, public authorities are trying their best to slow down the infection rate of the virus, globally. As part of their efforts, several solutions, such as closing public places, imposing full or partial lock-downs, and limiting people's contacts, have been implemented. Contact tracing has been globally recognized as one of the effective methods to slow down the infection rate of the virus \cite{kucharski2020effectiveness}. To this aim, most of the initial efforts were based on manually tracing the contacts of infected persons. Manual contact tracing works only when the infected person knows who has been in physical contact with him/her, which reduces the effectiveness of the method. Moreover, manual contact tracing is a very time- and resource-consuming process \cite{contact_tracing,rekanar2020sentiment}.   

The potential of contact-tracing could be fully utilized if, ideally, the contact tracing mechanism can track the contact of an infected person on a very large scale. For instance, it would be ideally beneficial if the authorities are able to track where the infected person has been and identify and notify the potential contacts of the patient. The technology, such as proximity sensors in smartphones and wearable devices, can help in such situations allowing the authorities to automatically notify the potential cases more quickly and accurately \cite{kucharski2020effectiveness,sinha2020contact}. To this aim, several mobile applications with a diversified set of features have been developed world-wide each aligned with COVID-19 related policies, social values, and local infrastructure. However, the success of such applications is largely constrained by the number of users. According to Hinch et al. \cite{hinch2020effective}, the potential of these mobile applications could be fully utilized if used by at least 80\% of the mobile users---which is 56\% of the total population in the case of UK as reported by the authors but generally depends on the mobile penetration in the country.

In order to increase the number of these applications' users, different strategies and policies have been devised \cite{shahroze2021}. For instance, public authorities in several countries have made it mandatory for the residents to install the contact tracing application to be able to access shopping malls, transportation, hospital, and other public places. 

However, there are several concerns over these applications in terms of both effectiveness and privacy. For instance, since the applications require tracking individuals' movement with GPS and other sensors to track their interactions, privacy concerns may arise \cite{abeler2020covid,de2020wetrace}. Moreover, the literature also identifies lack of understanding and unavailability of the technology (e.g., smartphones) with a large portion of the population in third world countries, as one of the main reasons for less effectiveness of such contact tracing applications \cite{altmann2020acceptability}. 

We believe an analysis of users' reviews on these applications will facilitate a better understanding of the concerns over these applications. There are already some efforts in this regard \cite{garousi2020mining,rekanar2020sentiment,altmann2020acceptability}. However, the majority of the methods rely on exploratory and manual analysis of the users' reviews, which is a resource and time-consuming process. Moreover, some of the works also rely on existing general sentiment analysis platforms/tools without training or fine-tuning the tools on COVID-19 application reviews. For instance, in \cite{garousi2020mining}, a commercial tool namely \textit{AppBot\footnote{https://appbot.co}} has been used for sentiment analysis of users' reviews on only nine mobile applications used in Europe. However, the tool relies on AI models trained for generic sentiment analysis, and returns four types of sentiments namely \textit{positive}, \textit{negative}, \textit{neutral}, and \textit{mixed}. As a result, the outcome is not reliable as the models are not trained on the task-specific data (i.e., App reviews). For instance, a vast majority of the reviews are highlighting some technical issues, such as difficulties with registration, etc., which also need to be analyzed. To address those limitations, we believe a task-specific model trained on a large-scale manually annotated users' reviews dataset will help to make better and context-specific classification of the reviews. Moreover, the existing literature relies on the users' reviews of fewer applications used in a specific region, which cover only a portion of the population of the world. 

In this paper, we analyze \textit{how AI models can help in automatically extract and classify the polarity of users' sentiments}, and propose a sentiment analysis framework to automatically analyze users' reviews on COVID-19 contact tracing mobile applications. In detail, we collected and annotated a large-scale dataset of Andriod and iOS mobile applications users' reviews for COVID-19 contact tracing. 
After manually analyzing and annotating users' reviews,
% in a crowd-sourcing study, 
we employed both classical (i.e., MNB, SVM, Random Forest) and deep learning (i.e., Convolution Neural Network \cite{krizhevsky2012imagenet}, fastText \cite{joulin2017bag}, and different transformers \cite{devlin2018bert}) methods for classification experiments. This resulted in eight different classification models.
% including a baseline.
% eight different methods including a BoW, BERT, Distil-BERT, Fasttext, ROBERTA, and XML-ROBERTA. In the conventional method, we employed BERT and Bag of Words (BoW) techniques while for an explanation of the predictions made by the automatic sentiment analyzer we rely on YYYY method. 
Moreover, to the best of our knowledge, this is the first attempt to develop a large-scale benchmark dataset for sentiment analysis of users' reviews on COVID-19 contact tracing applications, which are
% where applications 
from 46 distinct countries from \textit{Google Play} and \textit{App Store} are covered.

The main contributions of the work can be summarized as follows:
\textcolor{black}{
\begin{itemize}
    \item We provide \textit{34,534} manually labeled reviews based on the analysis of 40,000 reviews from 46 different COVID-19 contact tracing applications. The labels consist of sentiment polarities (i.e., positive, neutral, and negative) and a label (technical issue). 
    \item We provide an in-depth analysis of the dataset that demonstrates different characteristics and insights. 
    \item We share the dataset and data splits\footnote{Available at: \url{https://doi.org/10.7910/DVN/1RDRCM}} with the research community for both reproducibility and further enhancements.    
    \item We report benchmark results using eight different classification experiments, which can serve as a baseline for future studies. 
\end{itemize}
}

The rest of the paper is organized as follows: Section~\ref{sec:related_work} provides an overview of the related work. Section~\ref{sec:dataset} discusses the development of the dataset.
% describes the proposed methodology adopted for the sentiment analysis of users' reviews.
Section~\ref{sec:experiments} details the classification experiments.
% statistics of the crowd-sourcing study along with the 
The detailed observations of our study are discussed in Section~\ref{sec:discussion}.
% experimental results, and lessons learned from the experiments. 
Section~\ref{sec:conclusions} concludes this study and provides directions for future research. 

\section{Related Work}
\label{sec:related_work}
% Similar to other communities, the Data Science community is also playing its part in the battle against COVID-19. 
To fight against the COVID-19 pandemic, almost all research communities, such as health, NLP, and Computer Vision have been playing a significant role. As a result, several interesting solutions, aiming at different aspects of the pandemic, have been proposed over the last year \cite{latif2020leveraging}. For instance, there have been efforts for an early COVID-19 outbreak detection to help in an emergency response preparedness \cite{gharavi2020early}. Similarly, a large portion of the efforts aimed at an automatic diagnosis, prognosis, and treatment \cite{gharavi2020early,qayyum2021collaborative}. Fake news detection, risk assessment, logistic planning, and understanding of social interventions, such as monitoring social distancing, are the other key aspects of the pandemic that received the attention of the community \cite{hamid2020fake,bang2021model,latif2020leveraging}. 

Contact tracing is also one of the aspects of the pandemic that has been widely explored in the literature. For instance, the study by Lash et al. \cite{lash2020covid} analyzed the mechanism and results of contact tracing in two different countries. The authors report that an accurate and efficient mechanism of contact tracing can significantly reduce the infection rate of the virus.
% According to Lash et al. \cite{lash2020covid}, who analyzed the mechanism and results of contact tracing in two different countries, an accurate and efficient mechanism of contact tracing can significantly reduce the infection rate of the virus. 
However, several challenges are associated with a timely and accurate contact tracing of a COVID-19 patient. In this regard, a joint effort from the community, and the 
use of more advanced methods relying on different technologies, such as global positioning system (GPS), Wireless Fidelity (Wi-Fi), Bluetooth, Social graph, network-based APIs, and mobile tracking data will help to a great extent \cite{mbunge2020integrating,lalmuanawma2020applications}. Handheld devices, such as mobile phones, which are already embedded with such technologies, are ideal platforms for deploying contact tracing solutions. Being a feasible solution, several mobile applications have been already developed in different parts of the world.
% , where besides, 
In addition to basic contact tracing capabilities different features 
% in-lined with 
are also implemented based on the domestic COVID-19 policies \cite{ahmed2020survey}. 
For instance, in different countries, such as Qatar, Australia, the applications are used to access different facilities. Similarly, in Saudi Arabia, the application is used to seek permission for going out during lockdown. In Table \ref{tab:list_apps}, we provide a list of some of the prominent contact tracing applications used in different parts of the world.

Despite being a feasible solution for slowing down the infection rate, these applications are subject to criticism due to risks associated with them. In the literature, several issues, such as privacy, power consumption, and annoying alerts, have been reported. For instance, Bengio et al. \cite{bengio2021inherent} analyze and reported the privacy issues associated with COVID-19 contact tracing applications. Besides, some recommendations on how to ensure users' privacy, the authors also proposed a decentralized design for contact tracing by optimizing the privacy and utility trade-offs. Reichert et al. \cite{reichert2020privacy} also analyzed the privacy concerns over the applications, and proposed a privacy-preserving contact tracing mechanism relying on a privacy-preserving protocol namely secure multi-party computation (MPC) \cite{smart2016cryptography} to ensure individuals' privacy. Power consumption is another key challenge to contact tracing applications. 

The literature also provides several interesting works, where the feasibility of such mobile applications is carried out by analyzing people's response/feedback on these applications \cite{ahmed2020survey,li2020covid,cho2020contact}. For instance, in \cite{o2020national} an online survey was conducted to analyze citizens' response to HSE\footnote{\url{https://www.hse.ie/eng/}}, a contact tracing application used in Ireland. During the survey, a reasonable percentage of the participants showed their intention of using the application. However, the survey mainly aimed to analyze and identify different barriers in the use of such an application without analyzing the experience of the users with the application. In order to better analyze, understand, and evaluate users' experience and feedback on the COVID-19 contact tracing applications, a detailed analysis of the public reviews is required, which are available of Apple\footnote{https://www.apple.com/qa/app-store/} and Google Play\footnote{https://play.google.com/store} Store. 
%is required. 
There are already some efforts in this direction. For instance, Rekanar et al. \cite{rekanar2020sentiment} provide a detailed analysis of users' feedback on HSE in terms of usability, functional effectiveness, and performance. However, the authors rely on manual analysis only, which is a time-consuming process.  Another relevant work is reported in \cite{garousi2020mining}, where an exploratory analysis of users' feedback on nine COVID-19 contact tracing applications, used in Europe, is provided. To this aim, the authors rely on a commercial app-review analytics tool, namely Appbot\footnote{https://appbot.co}, to extract and mine the users' reviews on the applications. To the best of our knowledge, the literature still lacking a benchmark dataset to train and evaluate ML models for automatic analysis of users' feedback on COVID-19 contact tracing applications. Moreover, the existing literature relies on the users' reviews of fewer applications used in a specific region, which cover only a portion of the population of the world. \textcolor{black}{Hence, our work differs in a number of ways (i) we analyze reviews of a large number of applications used in different parts of the world, (ii) manually annotated dataset and provide them for the community, and (iii) provide detail experimental results.} 

 \begin{figure*}[]
     \centering
 	\includegraphics[width=0.89\textwidth]{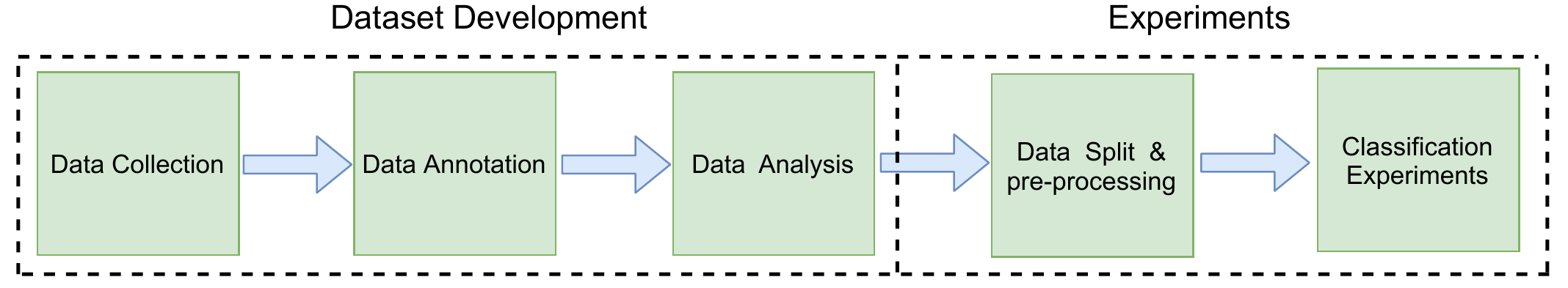}
 	\caption{Block diagram of the proposed pipeline for sentiment analysis of users' feedback on COVID-19 contact tracing mobile applications. The proposed pipeline can be roughly divided into two components, namely (i) dataset development, and (ii) experiments.}
 	\label{fig:methodology}
 \end{figure*}

% %%%%%%%%%%%%%%%%%%%%%%%%%%
% \section{Methodology}
\section{Proposed Pipeline}
\label{sec:methodology}
In this section, we will provide an overview of the methodology adopted in this paper. The complete pipeline of the proposed work is depicted in Figure \ref{fig:methodology}. The work is mainly carried out in two different phases including {\em (i)} dataset development phase, and {\em (ii)} experimental phase. In the development phase, as a first step, we scraped Google Play and App Store to obtain users' reviews on the COVID-19 contact tracing application used in different parts of the world. After obtaining users' reviews, a crowd-sourcing study has been conducted to annotate the reviews for the training and evaluation of ML models for automatic sentiment analysis of the reviews. The annotation and other details obtained during the crowd-sourcing study are then analyzed as detailed later. In the experimental phase, before conducting experiments, data is pre-processed to make it more meaningful for the AI models deployed in the experiments. During the experiments, we employed several AI models as detailed later. In the remainder of this section, we provide details of our dataset development process in which we describe our methodology for collecting, annotating, and analyzing the dataset.

%\section{Proposed Pipeline}
%\label{sec:pipeline}

%\fa{Let  us have the dataset section separate  to highlight its importance. 
%I think it is fine to have a section (III proposed pipeline) with a single paragraph. An example can be seen here section 2: https://arxiv.org/pdf/2007.07996.pdf}
\section{Dataset Development}
\label{sec:dataset}
% \subsection{Data Collection and Sentiment Categories Selection}
\subsection{Data Collection}
In order to obtain real-world users' reviews for our analysis, we crawled reviews from 46 COVID-19 contact tracing applications used in the different parts of the world and hosted on Google Play and Apple's App Store. These applications are listed in Table \ref{tab:list_apps}. 
% Google Play and App store are crawled  
% In total, we obtained users' reviews on 48 different applications as listed in Table \ref{tab:list_apps}. 
We note that in this work we consider reviews in the English language only where we made sure to analyze and annotate at least 50\% of reviews. However, to make sure the dataset is balanced in terms of reviews from different applications, for some applications, such as Aarogya Setu, a lesser portion of the available reviews is analyzed. Besides users' reviews, we also obtained replies to the reviews, if any are available, as well as the ratings. However, for this study, we only used the reviews for the analysis and experiments. We note that the reviews were obtained from December 20th to December 25th, 2020.

\begin{table}[]
\centering
\scalebox{0.75}{
\begin{tabular}{@{}llll@{}}
\toprule
\textbf{S. No.} & \textbf{Country} & \textbf{App} & \textbf{Tech.} \\ \midrule
1 & Australia & COVIDSafe & Bluetooth, Google/Apple \\
2 & Austria  & Stopp Corona & Bluetooth, Google/Apple\\
3 & Bahrain & BeAware & Bluetooth, Location \\
4 & Bangladesh & Corona Tracer BD & Bluetooth, Google\\ 
5 & Belgium & Coronalert & Bluetooth, Google/Apple\\
6 & Bulgaria & ViruSafe & Location, Bluetooth, Google/Apple \\
7 & Canada & COVID Alert & Bluetooth, Google/Apple \\
%9 & China & Chinese health code & Location, Data mining \\
8 & Cyprus &  CovTracer & Location,GPS \\
9 & Czech Republic  & eRouska & Bluetooth, Google/Apple \\
10 & Denmark & Smittestop  & Bluetooth, Google/Apple\\
11 & Estonia & HOIA & Bluetooth, DP-3T, Google/Apple\\
12 & Fiji & CareFiji & Bluetooth, Google/Apple \\
13 & Finland & Koronavilkku & Bluetooth, DP-3T \\
14 & France & TousAntiCovid  & Bluetooth, Google/Apple \\
15 & Germany & Corona-Warn-App & Bluetooth, Google/Apple\\
16 & Ghana & GH COVID-19 Tracker & Location, Google/Apple\\
17 & Gibraltar & Beat Covid Gibraltar & Bluetooth, Google/Apple \\
18 & Hungary  & VirusRadar & Bluetooth, Google \\
19 & Iceland & Rakning C-19 & Location, Google/Apple \\
20 & India & Aarogya Setu & Bluetooth, Location, Google/Apple\\
21 & Indonesia  & PeduliLindungi & Bluetooth, Google/Apple \\ 
%24 & Iran & AC-19 & Location  \\
22 & Ireland  & Covid Tracker & Bluetooth, Google/Apple \\
23 & Israel & HaMagen & Location, Google/Apple \\
24 & Italy & Immuni & Bluetooth, Google/Apple \\
25 & Japan & COCOA & Google/Apple  \\
26 & KSA & Tawakkalna & Bluettoth, Google\\
27 & KSA & Tabaud &  Google\\
28 & Kuwait & Shlonik & Location, Google/Apple \\
29 & Malaysia & MyTrace & Bluetooth, Google/Apple  \\
30 & Mexico & CovidRadar & Bluetooth\\ 
31 & New Zealand & NZ COVID Tracer & QR codes, Google/Apple \\
32 & North Macedonia  & StopKorona & Bluetooth \\
33 & Northern Ireland & StopCOVID NI & Bluetooth, Google/Apple\\
34 & Norway & Smittestopp & Bluetooth, Location, Google \\
35 & Pakistan & COVID-Gov-PK & Bluetooth, GPS, Google/Apple\\
36 & Philippines & StaySafe & Bluetooth, Google/Apple \\
37 & Poland & ProteGO Safe & Bluetooth, Google\\ 
38 & Qatar & Ehteraz & Bluetooth, Location, Google/Apple\\
39 & Singapore & TraceTogether & Bluetooth, Google/Apple\\
40 & South Africa  & COVID Alert SA & Bluetooth, Google/Apple \\
41 & Switzerland & SwissCovid & Bluetooth, DP-3T, Google/Apple \\
42 & Thailand & MorChana & Location, Bluetooth \\
43 & Tunisia & E7mi & Google/Apple \\
44 & Turkey & Hayat Eve Sığar & Bluetooth, Location, Google/Apple\\
45 & UAE & TraceCovid & Bluetooth \\
46 & UK & NHS COVID-19 App & Bluetooth, Google/Apple \\
 \bottomrule
\end{tabular}}
\caption{COVID-19 contact tracing mobile applications used in this study.}
\label{tab:list_apps}
\end{table}

%%%%%%%%%%%%%%%%%%%%%%%%%%%

% \subsection{Crowd-sourcing for Data Annotation}
\subsection{Data Annotation}
For the annotation of sentiment, typically three sentiment polarities are used such as \textit{positive}, \textit{negative}, and \textit{neutral}. From our initial analysis we realized that applications can have technical problems, hence, we used another label, \textit{technical issues}, for the annotation. Hence, our annotation consists of four labels: (i) \textit{positive}, (ii) \textit{negative}, (iii) \textit{neutral}, and (iv) \textit{technical issues}.

\begin{figure*}[!ht]
    \centering
	\includegraphics[width=0.75\textwidth]{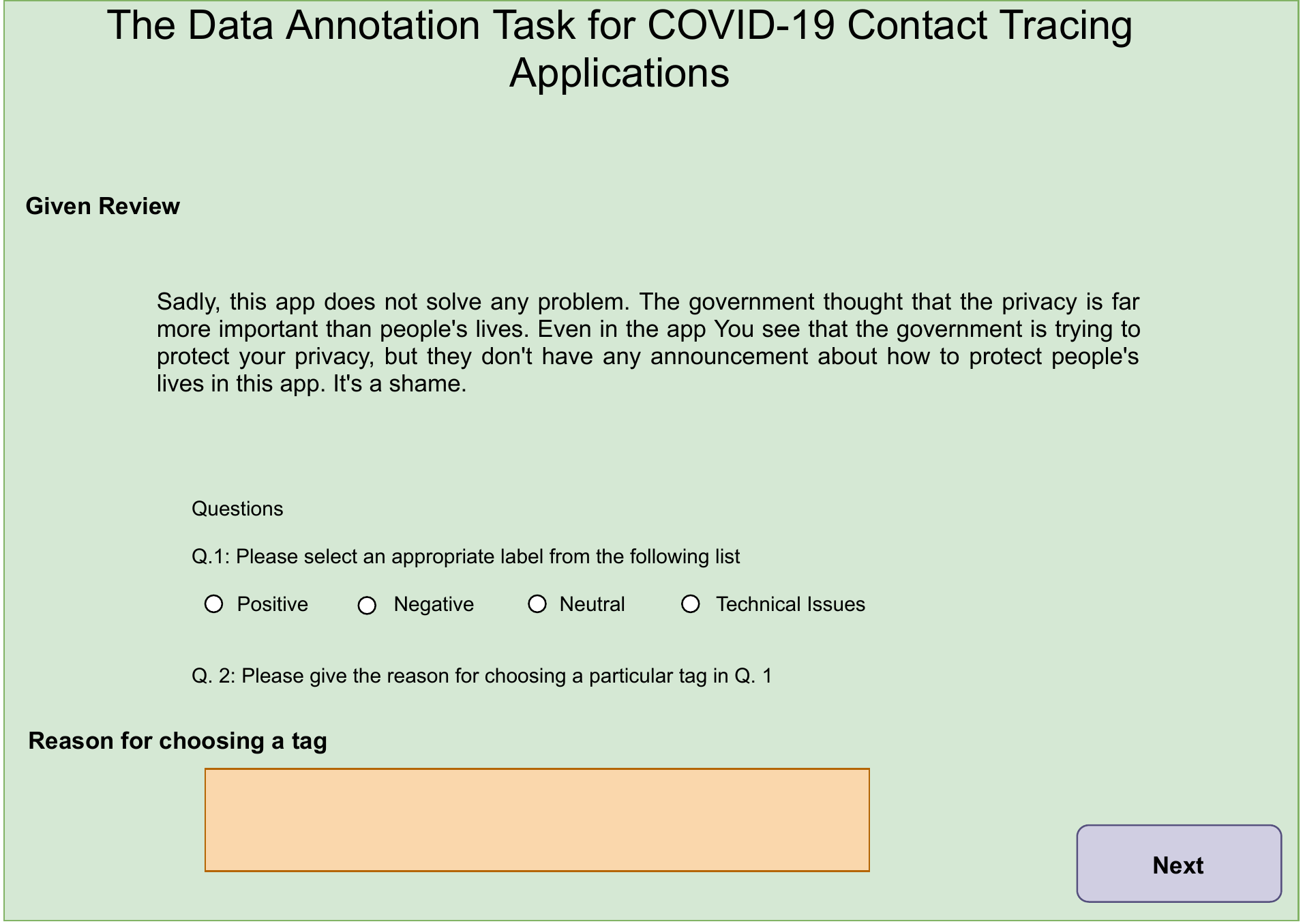}
	\caption{Block diagram of the annotation platform.}
	\label{fig:cs_study}
\end{figure*}

% The crowd-sourcing activity was conducted to develop ground-truth for the training and evaluation of the ML model used for the automatic sentiment analysis of the users' feedback. 
To facilitate the annotation process, we developed a web application
% To this aim, a web application was developed, 
where the users' reviews on the applications were presented to the annotators to be manually label them. % analyzed and annotated. 
In Figure \ref{fig:cs_study} we present a screen-shot of the annotation platform, which demonstrates the review and labels to be annotated (Q.1). In addition, we asked the annotators to briefly provide the reason behind their decision provided in response to Q.1. The question (Q.2) is used to evaluate the quality of the annotation (i.e., whether the annotator carefully read the review or not). Moreover, we believe this question will provide useful information for the manual analysis of the users' feedback.

In total, 40,000 reviews were analyzed. To assure the quality of the annotations, each review is analyzed by at least two participants (graduate students from different age groups). During the annotation process, we removed some reviews due to reasons such as (i) not being in English, (ii) having a large number of emoticons/signs, and (iii) they are irrelevant. This process resulted in having a total of \textit{\textbf{34,534}} annotated reviews. 
%with 15,587 reviews labeled positive; 8,178 reviews negative; 1,271 neutral; and 9,496 were labeled technical issues.

% In Figure \ref{fig:cs_study} provides a block diagram of the application used for the crowd-sourcing activity. As can be seen, users' feedback/reviews were presented to participants one by one, and a couple of questions were asked about the review. In the first question, participants were provided with four tags where they needed to choose a more appropriate one. In the second question, we asked the participants to briefly provide the reason behind their decision (i.e., the type of information that influenced their decision in question 1). This question is used to evaluate the quality of the annotation (i.e., whether the annotator carefully read the review or not). Moreover, we believe this question will provide useful information for the manual analysis of the users' feedback. 

% In total, around 40,000 reviews from Google Play and App Store were analyzed images during the activity. To assure the quality of the annotations, each review is analyzed by at least two participants. We note that participants included Masters and Ph.D. students from different age groups. 

%%%%%%%%%%%%%%%%%%%%%%%%%%%%%
%%%%%%%%%%%%%%%%%%%%%%%%%%

\subsection{Analysis}

\label{ssec:analysis}
% In total, the cleaned and annotated dataset is composed of 34,588 samples. 
%In Table~\ref{tab:dataset}, we provide number of annotated reviews per class. 
% provides the statistics of the dataset in terms of total number of samples per class. 
Overall, the dataset provides sufficient number of samples per class. However, fewer samples are obtained in the \textit{Neutral} category, which contains reviews that are not directly linked to the usage of the applications. In total, we have 15,587 samples in the \textit{positive} class while the \textit{negative}, \textit{neutral}, and \textit{technical issues} classes are composed of $8,178$, $1,271$, and $9,496$, respectively.

%%% TABLE DATASET %%%
% \begin{table}[]
% \caption{Statistics of the dataset.}
% \label{tab:dataset}
% \centering
% \begin{tabular}{|c|c|}
% \hline
% \textbf{Class labels} & \textbf{\# Reviews} \\ \hline
% Positive & 15,585  \\ \hline
% Negative & 8,215  \\ \hline
% Neutral & 1,274 \\ \hline
% Technical Issues & 9,514 \\ \hline
% \textbf{Total} & \textbf{34,588} \\ \hline
% \end{tabular}
% \end{table}

%\begin{table}[]
%\centering
%\begin{tabular}{@{}lr@{}}
%\toprule
%\multicolumn{1}{c}{\textbf{Class}} & \multicolumn{1}{c}{\textbf{\# Reviews}} \\ \midrule
%Positive & 15,587 \\
%Negative & 8,178 \\
%Neutral & 1,271 \\
%Technical Issues & 9,496 \\ \midrule
%\textbf{Total} & \textbf{34,534} \\ \bottomrule
%\end{tabular}
%\caption{Statistics of the dataset in terms of number of reviews in each class of the dataset.}
%\label{tab:dataset}
%\end{table}

\begin{figure}
    \centering
    \begin{subfigure}[bt]{0.35\textwidth}
        \centering
        \includegraphics[width=\textwidth]{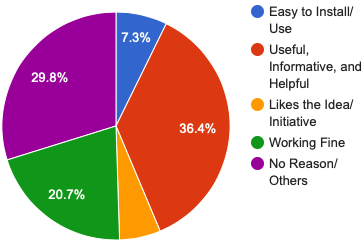}
        \caption{Causes for positive feedback}
        \label{fig:positive_stats}
    \end{subfigure}
    \hfill
    \begin{subfigure}[bt]{0.35\textwidth}
        \centering
        \includegraphics[width=\textwidth]{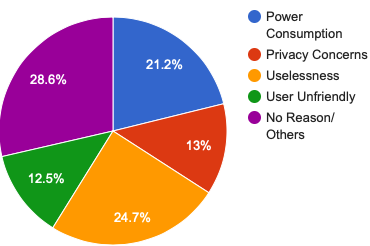}
        \caption{Causes for negative feedback}
        \label{fig:negative_stats}
    \end{subfigure}
    \begin{subfigure}[bt]{0.35\textwidth}
        \centering
        \includegraphics[width=\textwidth]{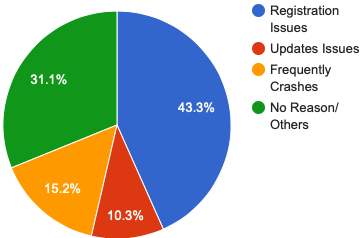}
        \caption{Causes for technical issues}
        \label{fig:technical_stats}
    \end{subfigure}
    \caption{Analysis of common reasons provided in the reviews.}
\end{figure}

From the analysis of the second question (Q.2), we identified the reasons/information that influenced the participants' decision. 
In this section, we provide the statistics of the second question (Q.2) 
% of the crowd-sourcing study, 
% where the participants were asked to identify the reason/information that influenced their decision. These statistics help to identify the most common issues with the COVID-19 contact tracing applications. 
In Figure \ref{fig:positive_stats}, \ref{fig:negative_stats}, and  \ref{fig:technical_stats}, we provide the distribution of the most common reasons/causes associated with the \textit{positive}, \textit{negative} and \textit{technical issues}, respectively. 

As can be seen in Figure \ref{fig:positive_stats}, in the majority of the positive reviews, users' found the applications useful, informative, and helpful in the battle against COVID-19. Some sample positive reviews are ``\textit{Thank you very much. it's very helpful and informative. it helps keep people away from suspicious areas.}''; ``\textit{A very good app for tracing and stopping coronavirus}''; ``\textit{Always getting updated information about the virus}''; ``\textit{Very useful and informative app.}'' A significant portion of positive reviews is also based on ease in the installation while some reviews mentioned that the application they are using is working fine without further details. However, the most encouraging aspect is the fact that a significant percentage of users have appreciated the idea, concept, and efforts made by the authorities for contact tracing to slow down the infection rate. Some sample reviews include ``\textit{A good initiative by the government}''; ``\textit{Good initiative to prevent the spread of corona virus, I appreciate who work behind this effort.}" There were also a large number of short reviews where the users simply showed their positive response without mentioning any particular reason. Besides these, other common reasons for their positive reviews highlighted by the users include some specific features of different applications in different parts of the world. For example, \textit{Takkawalna} from Saudi Government was used to seek permission for going out during lockdown was praised for being a source of seeking permissions.

Figure \ref{fig:negative_stats} reflects power consumption, uselessness, and privacy as the most common issues with these applications. A significant amount of reviews also highlight that majority of the applications are not user-friendly. There were also reviews depicting other issues, such as annoying notifications, unnecessary access to the gallery, slow response from the helpline, and unavailability of some key features, which could further improve the effectiveness of the applications. Some sample negative reviews include ``\textit{Too much personal information collected. Privacy risk. Non compliant to international standards.}"; ``\textit{Allow too many permission please ban this application. A total waste.}"; ``\textit{I have concerns with their data privacy.}"; and ``\textit{This app is a battery hog}."

On the other hand, as can be noticed in Figure \ref{fig:technical_stats}, the key technical issues with these applications include registration and update issues. Moreover, a large number of reviews also highlight that majority of the applications crashes or frequently stops working. Besides these common issues, the reviews also hint about certain technical issues, such as device compatibility and connectivity issues, lack of support for some languages, such as English, and not correcting QR codes by different applications. Some sample reviews highlighting technical issues in the applications include ``\textit{The app continually crashes.}"; ``\textit{I have business visa i am unable to register please give a solution on this.}"; ``\textit{I installed..but i cant register yet.}"'; ``\textit{I cant update?}"; and ``\textit{Install the apps, but keep showing connection error. Even restart the phone, also the same.}"

% %%%%%%%%%%%%%%%%%%%%%%%%%%%
% \begin{figure}[!htb]
% \minipage{0.5\textwidth}
%   \includegraphics[width=\linewidth]{img/pie_chart_positive.png}
%   \caption{Statistics of the most commonly reasons for positive reviews.}\label{fig:positive_stats}
% \endminipage\hfill
% \minipage{0.5\textwidth}
%   \includegraphics[width=\linewidth]{img/pie_chart_negative.png}
%   \caption{Statistics of the most commonly reasons for negative reviews.}\label{fig:negative_stats}
% \endminipage\hfill
% \minipage{0.5\textwidth}%
%   \includegraphics[width=\linewidth]{img/pie_chart_technical.png}
%   \caption{Statistics of the most commonly reasons for technical issues.}\label{fig:technical_stats}
% \endminipage
% \end{figure}

\begin{figure*}[!ht]
    \centering
	\includegraphics[width=0.95\textwidth]{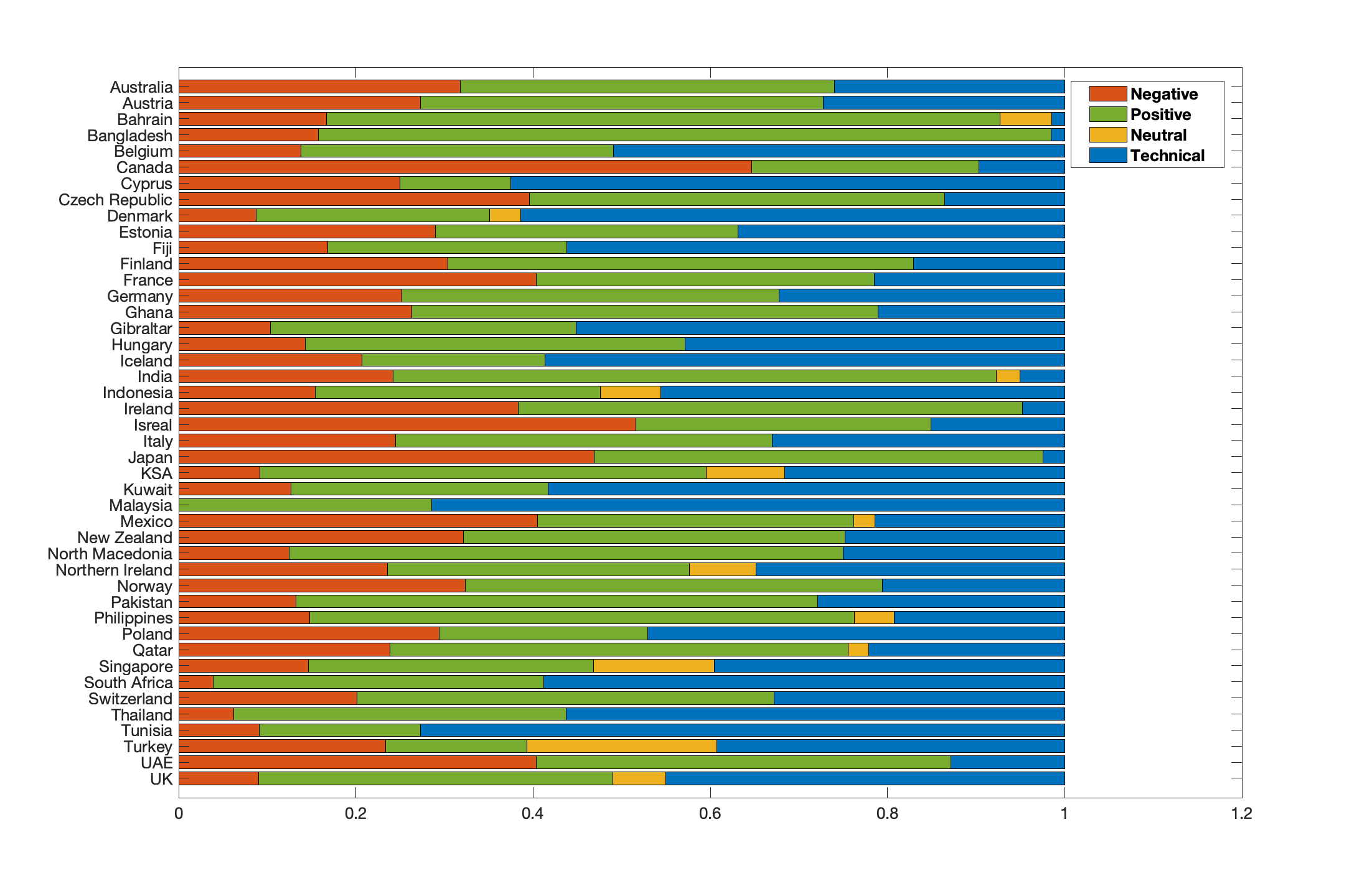}
	\caption{Country-wise statistics of the dataset in terms of the distribution of \textit{negative}, \textit{positive}, \textit{neutral}, and \textit{technical issues} in the applications covered in this work.}
	\label{fig:cs_study_country_wise}
\end{figure*}

We also provide country-wise statistics in Figure \ref{fig:cs_study_country_wise}, where we summarize the number/percentage of samples/reviews on the applications used in different countries belonging to each category.  %As can be seen, the dataset covers sufficient samples/reviews from different parts of the world, which is expected to help the trained models to be applicable anywhere. 
An important observation from the figure is the variation in the distribution of number of \textit{negative}, \textit{positive}, \textit{neutral}, and reviews highlighting \textit{technical issues} in different parts of the world. The variations in the number of reviews in each class depict how different response to the use of the applications has been observed in different parts of the world. As can be seen, in certain countries, such as Japan, Israel, Canada, and Ireland, the ratio of \textit{negative} reviews is high. The ratio of the \textit{positive} reviews is sufficient in most of the countries, which shows the trust of users in the applications. On the other hand, as expected, fewer \textit{neutral} reviews from the majority of the countries are obtained for the dataset. The dataset also covers a significant ratio for the \textit{technical issues} class in the majority of the countries. For instance, the ratio of the reviews highlighting technical problems in the applications is significantly high in \textit{Denmark}, \textit{Tunisia} and \textit{Cyprus}.

%%%%%%%%%%%%%%%%%%%%%%%%%%%%%
%%%%%%%%%%%%%%%%%%%%%%%%%%

\begin{figure*}[!ht]
    \centering
	\includegraphics[width=1\textwidth]{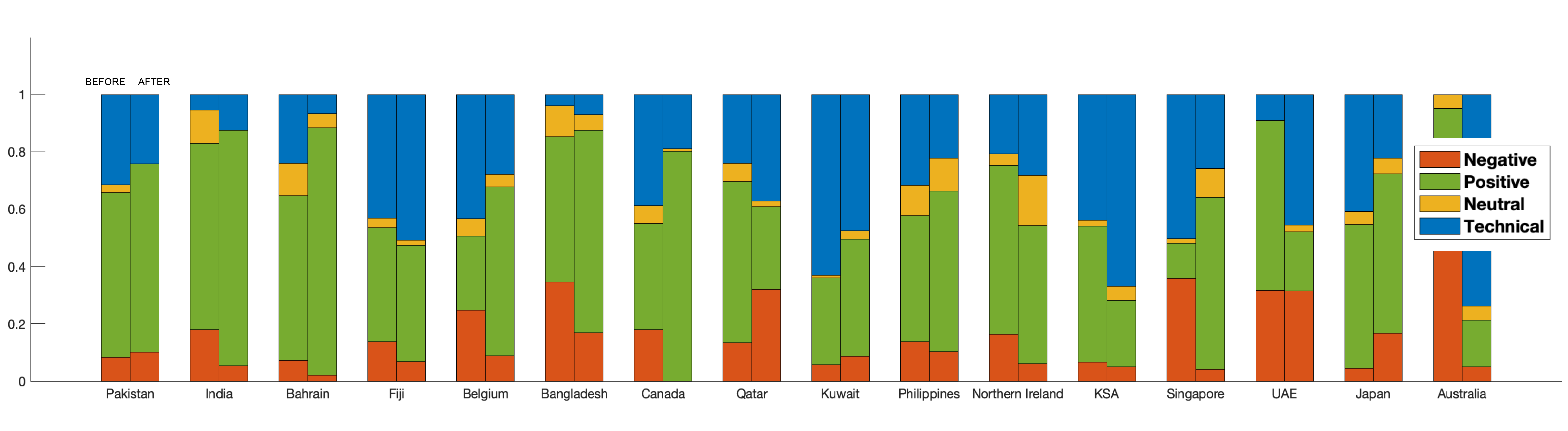}
	\caption{Preliminary temporal analysis reflecting the changes in the distribution of the sentiment classes over time. The statistics are compiled by analyzing the top 200 more recent (i.e., December 25th, 2020) and the initial 200 reviews from some applications having sufficient number of reviews.}
	\label{fig:cs_study_std}
\end{figure*}

In order to analyze the changes in the polarity of users' sentiments over time, in Figure \ref{fig:cs_study_std}, we provide some preliminary temporal analysis to analyze the variation in the distribution of \textit{negative}, \textit{positive}, \textit{neutral}, and reviews highlighting the \textit{technical issues} over time. We note that in the current work we provide some preliminary temporal analysis, which will be explored in the future, and to this aim, we manually analyzed the 200 most recent reviews and the initial 200 reviews on applications having a reasonable time duration in the initial and more recent reviews. As can be seen in the figure, overall higher variation has been observed in the \textit{positive}, \text{negative}, and \textit{neutral} categories. As far as the individual applications are concerned, higher variation in the polarity of sentiments is observed for the applications used in  \textit{Australia}, \textit{Singapore}, \textit{UAE}, and \textit{Canada}. 

During the data analysis, though there were some doubts about privacy, we observed that at the beginning the initiative/idea of contact tracing was largely appreciated by the users in different parts of the world. Moreover, we also observed that the users of these applications faced device compatibility and registration issues with the application with time. Interestingly, in the case of most of the applications, the number of \textit{negative} reviews increased with time. One of the possible reasons for the increase is the applications' failure in achieving what they promised. 

%%%%%%%%%%%%%%%%%%%%%%%%%%%%%
%%%%%%%%%%%%%%%%%%%%%%%%%%
\subsection{Lexical Analysis}
\label{ssec:lexical_analysis}

\begin{figure}[h]
\centering
\includegraphics[width=1.0\linewidth]{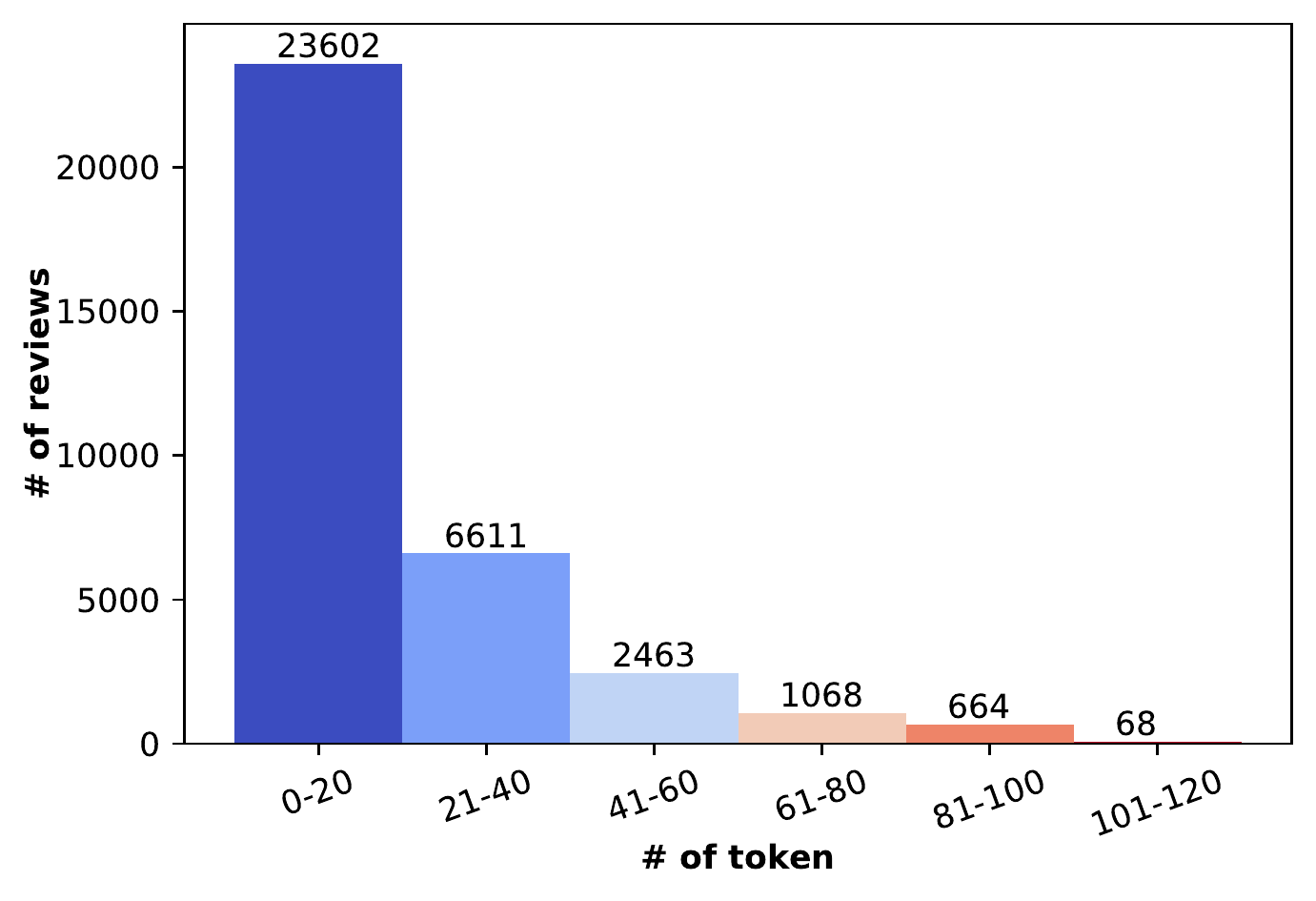}
\caption{Number of reviews with different lengths in the overall dataset. On 60 reviews we have more than 120 tokens, which has not shown in the figure.}
\label{fig:reviews_len_stat}
% \vspace{-0.3cm}
\end{figure}

To understand the lexical content, we have conducted an analysis of the number of tokens for each review. It can help to understand the characteristics of the dataset. For example, for CNN and LSTM based architectures, it is necessary to define max sequence length. The minimum, maximum and average number of tokens in the dataset are 3, 198, and $\sim$18, respectively. Figure \ref{fig:reviews_len_stat} provides the statistics of the length of the reviews in the dataset.

We also analyzed the lexical content in each category to understand whether they are distinctive in terms of the lexical content -- top n-grams. This analysis also demonstrates the quality of the labeled data. We compared the vocabularies of all categories using the valence score \cite{conover2011political,chowdhury2020multi}, $\vartheta$ for every token, $x$, using the following the Equation \ref{eq:valence}: 
% \vspace{-0.2em}	
\begin{equation}
\label{eq:valence}
    \vartheta (x, L_i) = 2 * \frac{\frac{C(x|L_i)}{T_{L_i}}}{\sum_{l}^{L} C(x|L_l) } -1
\end{equation}
\noindent where $C(.)$ is the frequency of the token $x$ for a given class $L_i$. $T_{L_i}$ is the total number of tokens present in the class. In $\vartheta(x) \in [-1, +1]$, the value $+1$ indicates the use of the token is significantly higher in the target class than the other classes.

In Table \ref{tab:class_wise_ngrams}, we present top frequent bi- and tri-grams with $\vartheta=1.0$ for each category. From the table, we observe these n-grams clearly represent the class-wise information of the data.

\begin{table}[h]
\centering
\scalebox{0.90}{
\begin{tabular}{@{}lll@{}}
\toprule
\multicolumn{1}{c}{\textbf{Negative}} & \multicolumn{1}{c}{\textbf{Positive}} & \multicolumn{1}{c}{\textbf{Technical issues}} \\ \midrule
Battery went down & Best app & Error requesting \\
to delete this & Excellent app, & Cannot register. \\
overheating and battery & and helpfull & what's wrong \\
Not happy & Very nice and & fix this, \\
battery drainer & feel safer & I can't seem \\
uninstall due to & very good apps & Unable to proceed \\
Drains battery and & Good information & error while \\
massive drain & save lives and & have error \\
too much battery, & very useful for & phone number. Tried \\
Massive battery drain & Easy to register & I force stop \\ \bottomrule
\end{tabular}
}
\caption{Class-wise n-grams based on valance scores.}
\label{tab:class_wise_ngrams}
\end{table}
% \section{Experiments and Evaluation}
\section{Experiments}
\label{sec:experiments}

%%%%%%%%%%%%%%%%%%%%%

\subsection{Task Description}
\label{ssec:task_description}
As discussed earlier, we obtained a large number of samples for \textit{positive}, \textit{negative}, and \textit{technical issues} while fewer samples are obtained in \textit{neutral} class. Moreover, the reviews highlighting technical problems in the applications could also be treated as negative reviews. Thus, in order to cover different aspects of the problem, we divide it into three different tasks.

\textbf{Task 1: Ternary classification (PNT)}---we treat the problem as ternary classification problem, where \textit{positive}, \textit{negative}, and \textit{technical issues} are considered. The models trained for this task are expected to help in identifying the reviews highlighting technical problems in the applications along with the positive and negative reviews. 

\textbf{Task 2: Binary classification (PN)}---the \textit{negative} and \textit{technical issues} classes are merged into a single \textit{negative} class to form two classes for a binary classification problem along with \textit{positive} reviews. One of the main reasons for treating the task as a binary classification is the availability of fewer samples in the \textit{neutral} class. 

% However, 
\textbf{Task 3: Ternary classification (PNN)}---we have three classes namely \textit{positive}, \textit{negative}, and \textit{neutral}. We note that in this task, the \textit{negative} class is the combination of original negative and technical issues classes. 
% In the third task, we also consider the \textit{neutral} class along with the \textit{positive}, and the newly formed \textit{negative} class resulted in the combination of the original \textit{negative} and \textit{technical issues} classes. 

All these tasks will help in analyzing how the performances of the proposed sentiment analyzer vary with different sets of annotations. 

\subsection{Data Splits}
\label{ssec:data_split}
For the classification experiments, we divided the dataset into training, validation, and test sets with a proportion of 60.3\%, 6.7\%, and 30\%, respectively. While dividing the dataset we used stratified sampling to maintain class distribution across different sets. 
% To this aim a python function has been used to automatically split the data. 
% \textcolor{blue}{Moreover, 
The data split/distribution is performed for each task separately, which results in a different number of samples for training, validation, and test set for each task. The data split for each task will be made publicly available, separately, to ensure a fair comparison in future work.
% } 
Table \ref{tab:data_split} summarizes the distribution of the data into training, validation, and test sets used in each task, respectively. 

\begin{table}[h]
\centering
\begin{tabular}{@{}lrrrr@{}}
\toprule
\multicolumn{1}{c}{\textbf{Class labels}} & \multicolumn{1}{c}{\textbf{Train}} & \multicolumn{1}{c}{\textbf{Dev}} & \multicolumn{1}{c}{\textbf{Test}} & \multicolumn{1}{c}{\textbf{Total}} \\ \midrule
\multicolumn{5}{c}{\textbf{Task 1: Ternary classification (PNT)}} \\ \midrule
Positive & 9370 & 1041 & 5176 & 15587 \\
Negative & 5000 & 556 & 2622 & 8178 \\
Technical Issues & 5686 & 632 & 3178 & 9496 \\ \cmidrule{2-5}
\textbf{Total} & 20056 & 2229 & 10976 & 33261 \\\midrule
\multicolumn{5}{c}{\textbf{Task 2: Binary classification (PN)}} \\\midrule
Positive & 9342 & 1038 & 5207 & 15587 \\
Negative & 10715 & 1191 & 5770 & 17676 \\ \cmidrule{2-5}
\textbf{Total} & 20057 & 2229 & 10977 & 33263 \\\midrule
\multicolumn{5}{c}{\textbf{Task 3: Ternary classification (PNN)}} \\\midrule
Positive & 9364 & 1040 & 5183 & 15587 \\
Negative & 10690 & 1188 & 5798 & 17676 \\
Neutral & 770 & 85 & 416 & 1271 \\ \cmidrule{2-5}
\textbf{Total} & 20824 & 2314 & 11398 & 34534 \\ \bottomrule
\end{tabular}
\caption{Data split and distribution of class labels for different tasks.}
\label{tab:data_split}
\end{table}

%%%%%%%%%%%%%%%%%%%%%%
%\begin{table*}
%\centering
%\begin{tabular}{|l|c|c|l|c|c|l|l|c|c|l|l|}
%\hline
%\multicolumn{4}{|c|}{\textbf{Task 1}} & \multicolumn{4}{c|}{\textbf{Task 2}} & \multicolumn{4}{c|}{\textbf{Task 3}} \\ \hline
%Label & Train & Test & Validation & Label & Train & Test & Validation & Label & Train & Test & Validation \\ \hline
% Positive&  &  &  & Positive &  &  &  &  &  &  &  \\ \hline
% Negative &  &  &  & Positive &  &  &  &  &  &  &  \\ \hline
%Technical &  &  &  &  &  &  &  &  &  &  &  \\ \hline
%\end{tabular}
%\end{table*}
%%%%%%%%%%%%%%%%%%%%
\subsection{Data Prepossessing}
% \todo[inline]{@Kashif, did you do any cleaning on the data, like removing any unnecessary things?} \textcolor{blue}{yes, we removed some reviews as whole as well as some unnecessary text e.g., /\ etc., and emoji and smilies etc., etc., }

Before proceeding with the experiments, the data is also cleaned by removing unnecessary tokens, such as non-ASCII characters, punctuations (replaced with whitespace), and other signs.
%%%%%%%%%%%%%%%%%%%%%%%%%%%%%
%\begin{figure*}
 %   \centering
%	\includegraphics[width=0.99\textwidth]{img/sentiment_analyzer_methodology.pdf}
%	\caption{Block diagram of the proposed methodology for the sentiment analysis.}
%	\label{fig:sentiment_anazlyser}
%\end{figure*}
%%%%%%%%%%%%%%%%%%%%%%%%%%
\subsection{Models}
For this study, our classification experiments consist of multiclass classification using both classical and deep learning algorithms as detailed below.

% \subsubsection{Bag of Words Model}
\subsubsection{Classical Algorithms}
For this study, we used several classical algorithms such as Multinomial Naive Bayes (MNB) \cite{Zhang2004TheOO}, SVM \cite{hearst1998support} and Random Forest (RF)~\cite{breiman2001random}. As a feature representation with these algorithms, we used the bag-of-ngrams, which is one of the most commonly used methods for text classification and retrieval applications, applied with classical algorithms. Earlier this has been widely used as a simple, yet effective and computationally efficient method. Motivated by its better performance in similar types of text classification applications, such as fake news and floods detection in Twitter text \cite{said2019natural,hamid2020fake,faustini2020fake}, we experimented with this representation using the mentioned classical algorithms.

\subsubsection{fastText}
fastText is an NLP library aiming at efficient word embedding and text classification with a higher speed compared to traditional deep learning solutions \cite{joulin2016bag}. For word embedding, the model relies on Continuous Bag of Words (CBOW), which is based on a shallow Neural Network (NN), strategy by predicting a word via its neighbors. To ensure training at a higher speed, the model relies on a hierarchical classification mechanism by replacing the traditional soft-max function with a hierarchical one resulting in a reduced number of parameters.

\subsubsection{Transformers}
BERT \cite{devlin2018bert} is a state-of-the-art pre-trained model, which has shown its success in many downstream NLP tasks. It is typically used for downstream classification problems either by using embedding representations as features or fine-tuning the model. The main strength of the model comes from pre-training on a very large text dataset that allows the model to understand and interpret text easily in different NLP applications. Moreover, the model also possesses the ability to learn from context. For this study, we use different transformer models, which  include BERT \cite{devlin2018bert}, RoBERTa \cite{liu2019roberta}, XLM-RoBERTa~\cite{conneau2019unsupervised} and DistilBERT \cite{sanh2019distilbert}. 

\subsection{Evaluation Measure}
To measure the performance of each classifier, we use weighted average precision (P), recall (R), F1-measure (F1). We used weighted metric as it has the capability to take into account the class imbalance distribution.

\subsection{Classification Experiments}
\label{ssec:classificaiton_exp}

To train the classifiers using the MNB, SVM, and RF we converted the text into bag-of-$n$-gram vectors weighted with logarithmic term frequencies (tf) multiplied with inverse document frequencies (idf). To utilize contextual information, such as $n$-grams which are useful for classification, we extracted unigram, bigram, and tri-gram features. 

We used grid-search to optimize the parameters for MNB, SVM, and RF. For the MNB, we optimize laplace smoothing $\alpha$ parameter with 20 values between 0 and 1. For the SVM, we optimize linear kernel with $C$ parameters with 30 values ranges from 0.00001 to 10, and radial-basis-function kernel with $C$ and $\gamma$ parameters (for $\gamma$ we use 10 values from 1e-5 to 1e-1). For RF we optimize the number of trees (10 values from 200 to 2000), and the depth of the tree (11 values from 10 to 110). Choosing such ranges of values depends on the available computational resources as they are computationally expensive.

% For SVM experiments, we applied grid search on the following parameters: RBF and linear kernel, $C$ and $\gamma$ parameters. For RF experiments, we set the number of trees to 100.

For fastText, we use pre-trained embeddings trained on Common Crawl\footnote{\url{https://fasttext.cc/docs/en/crawl-vectors.html}} and default hyperparameter settings available with fastText toolkit.\footnote{\url{https://fasttext.cc/}}

For transformer-based models, we use the Transformer Toolkit~\cite{Wolf2019HuggingFacesTS}. We fine-tune each model using the below hyper-parameter settings with a task-specific layer on top of the model. As reported in \cite{devlin2018bert} the training with the pre-trained transformer models shows instability, hence, we do 10 runs of each experiment using different random seeds and choose the model that performs the best on the development set. For training the transformer-based models for each task we fine-tune the model 10 epochs with the `categorical cross-entropy' as the loss function and used the following hyper-parameter settings.
\begin{itemize}
    \itemsep0em
    \item Batch size: 8
    \item Learning rate (Adam): 2e-5
    \item Number of epochs: 10
    \item Max seq length: 128
    % \item Weight Decay: 0.0
    % \item Learning Rate Decay:
    % \item Warmup ratio: 
\end{itemize}

Below we provide the detail of the number of parameters\footnote{\url{https://huggingface.co/transformers/pretrained_models.html}} for each model, which demonstrates the size of the models. 
\begin{itemize}
    \itemsep0em
    \item \textbf{BERT} (bert-base-uncased): This model is trained on lower-cased English text. It consists of 12-layer, 768-hidden, 12-heads, 110M parameters.
    \item \textbf{DistilBERT} (distilbert-base-uncased): This is a distilled version of the BERT model consists of 6-layer, 768-hidden, 12-heads, 66M parameters.    
    \item \textbf{RoBERTa} (roberta-large): RoBERTa using the BERT-large architecture consists of 24-layer, 1024-hidden, 16-heads, 355M parameters.
    \item \textbf{XML-RoBERTa} (xlm-roberta-large): It consists of $\sim$355M parameters with 24-layers, 1027-hidden-state, 4096 feed-forward hidden-state, and 16-heads.
\end{itemize}

\section{Results}
\label{sec:results}

% \subsection{Experimental Results on Task 1}
\subsection{Task 1: Ternary classification (PNT)}
Table \ref{tab:result_task1} provides the experimental results on task 1 in terms of weighted accuracy, precision, recall, and F1-Score. Overall better results are obtained with transformers compared to the classical and deep learning based methods. One of the main reasons for the better performance of the transformers is their text interpretation capabilities. Though no significant differences have been observed in the performances of the different transformers, a slight improvement is observed for RoBERTa over the rest of transformers. 

To better analyze the performance of the proposed methods, we also provide class-wise performance. Overall reasonable results are obtained on all three classes, however, the performance of all the methods is higher on the \textit{positive} class. One of the possible reasons for the comparatively lower performance on the other two classes is the lower inter-class variation. As detailed earlier, reviews in the \textit{negative} and \textit{technical issues} classes contain similar types of words, and there are higher chances of confusion in the classes. The experimental results of task 1 provide bases for task 2, where the \textit{negative} and \textit{technical issues} classes are merged. 

\begin{table*}[h]
\centering
\begin{tabular}{@{}lrrrrrrrrrrrrr@{}}
\toprule
\multicolumn{1}{c}{\textbf{}} & \multicolumn{3}{c}{\textbf{Positive}} & \multicolumn{3}{c}{\textbf{Negative}} & \multicolumn{3}{c}{\textbf{Technical Issues}} & \multicolumn{4}{c}{\textbf{Overall (Weighted Average)}} \\ \midrule
\multicolumn{1}{c}{Method} & \multicolumn{1}{c}{P} & \multicolumn{1}{c}{R} & \multicolumn{1}{c}{F1} & \multicolumn{1}{c}{P} & \multicolumn{1}{c}{R} & \multicolumn{1}{c}{F1} & \multicolumn{1}{c}{P} & \multicolumn{1}{c}{R} & \multicolumn{1}{c}{F1} & \multicolumn{1}{c}{Acc} & \multicolumn{1}{c}{P} & \multicolumn{1}{c}{R} & \multicolumn{1}{c}{F1} \\ \midrule
MNB & 0.910 & 0.892 & 0.901 & 0.679 & 0.664 & 0.671 & 0.751 & 0.789 & 0.769 & 0.808 & 0.809 & 0.808 & 0.808 \\
RF & 0.854 & 0.923 & 0.887 & 0.809 & 0.538 & 0.646 & 0.729 & 0.833 & 0.777 & 0.805 & 0.806 & 0.805 & 0.797 \\
SVM & 0.946 & 0.867 & 0.905 & 0.660 & 0.707 & 0.683 & 0.745 & 0.803 & 0.773 & 0.810 & 0.820 & 0.810 & 0.814 \\
fastText & 0.930 & 0.904 & 0.917 & 0.713 & 0.691 & 0.702 & 0.752 & 0.806 & 0.778 & 0.825 & 0.827 & 0.825 & 0.825 \\
DistilBERT & 0.943 & 0.934 & 0.939 & 0.753 & 0.714 & 0.733 & 0.778 & 0.824 & 0.800 & 0.849 & 0.850 & 0.849 & 0.849 \\
BERT & 0.938 & 0.936 & 0.937 & 0.750 & 0.718 & 0.734 & 0.786 & 0.817 & 0.801 & 0.850 & 0.849 & 0.850 & 0.849 \\
RoBERTa & 0.943 & 0.946 & 0.945 & 0.754 & 0.716 & 0.734 & 0.788 & 0.817 & 0.802 & 0.854 & 0.853 & 0.854 & 0.853 \\
XML-RoBERTa & 0.941 & 0.946 & 0.943 & 0.744 & 0.705 & 0.724 & 0.783 & 0.811 & 0.797 & 0.849 & 0.848 & 0.849 & 0.848 \\ \bottomrule
\end{tabular}
\caption{Experimental results on \textbf{Task 1: Ternary classification (PNT)}. 
% \textcolor{red}{some mismatch in combining the tables-to be checked yet}
}
\label{tab:result_task1}
\end{table*}
%%%%%%%%%%%%%%%%%%%%%%%%

\subsection{Task 2: Binary classification (PN)}
Table \ref{tab:result_task2} provides experimental results on task 2, where the models have to differentiate between \text{positive} and \textit{negative} reviews. As expected, the performance has been improved significantly on task 2, which proves our hypothesis that \textit{negative} and \textit{technical issues} classes have similarity in contents. Moreover,  similar to task 1, transformers have outperformed the rest of the methods. 

As can be seen in the table, in contrast to task 1, no significant differences can be observed in the performance of the methods on different classes, which indicates that reviews highlighting technical problems in the applications evoke negative emotions/sentiments. Moreover, no significant variation in the performance of the methods on a particular class has been observed.

\begin{table*}[h]
\centering
\begin{tabular}{@{}lrrrrrrrrrr@{}}
\toprule
\multicolumn{1}{c}{\textbf{}} & \multicolumn{3}{c}{\textbf{Positive}} & \multicolumn{3}{c}{\textbf{Negative}} & \multicolumn{4}{c}{\textbf{Overall (Weighted Average)}} \\ \midrule
\multicolumn{1}{c}{\textbf{Method}} & \multicolumn{1}{c}{P} & \multicolumn{1}{c}{\textbf{R}} & \multicolumn{1}{c}{F1} & \multicolumn{1}{c}{\textbf{P}} & \multicolumn{1}{c}{R} & \multicolumn{1}{c}{\textbf{F1}} & \multicolumn{1}{c}{Acc} & \multicolumn{1}{c}{\textbf{P}} & \multicolumn{1}{c}{R} & \multicolumn{1}{c}{\textbf{F1}} \\ \midrule
MNB & 0.925 & 0.873 & 0.898 & 0.891 & 0.936 & 0.913 & 0.906 & 0.907 & 0.906 & 0.906 \\
RF & 0.902 & 0.879 & 0.891 & 0.894 & 0.914 & 0.904 & 0.898 & 0.898 & 0.898 & 0.898 \\
SVM & 0.944 & 0.876 & 0.909 & 0.895 & 0.953 & 0.923 & 0.916 & 0.918 & 0.916 & 0.916 \\
fastText & 0.947 & 0.890 & 0.917 & 0.905 & 0.955 & 0.929 & 0.924 & 0.925 & 0.924 & 0.924 \\
DistilBERT & 0.947 & 0.932 & 0.939 & 0.939 & 0.953 & 0.946 & 0.943 & 0.943 & 0.943 & 0.943 \\
BERT & 0.947 & 0.936 & 0.941 & 0.943 & 0.953 & 0.948 & 0.945 & 0.945 & 0.945 & 0.945 \\
RoBERTa & 0.948 & 0.942 & 0.945 & 0.948 & 0.953 & 0.951 & 0.948 & 0.948 & 0.948 & 0.948 \\
XML-RoBERTa & 0.953 & 0.930 & 0.942 & 0.939 & 0.959 & 0.949 & 0.945 & 0.946 & 0.945 & 0.945 \\ \bottomrule
\end{tabular}
\caption{Experimental results on \textbf{Task 2: Binary classification (PN)}. 
% \textcolor{red}{some mismatch in combining the tables-to be checked yet}
}
\label{tab:result_task2}
\end{table*}
%%%%%%%%%%%%%%%%%%%%%%%

%%%%%%%%%%%%%%%%%%%%%%%%

\begin{table*}[h]
\centering
\begin{tabular}{@{}lrrrrrrrrrrrrr@{}}
\toprule
\multicolumn{1}{c}{\textbf{}} & \multicolumn{3}{c}{\textbf{Positive}} & \multicolumn{3}{c}{\textbf{Negative}} & \multicolumn{3}{c}{\textbf{Neutral}} & \multicolumn{4}{c}{\textbf{Overall (Weighted Average)}} \\ \midrule
\multicolumn{1}{c}{Method} & \multicolumn{1}{c}{P} & \multicolumn{1}{c}{R} & \multicolumn{1}{c}{F1} & \multicolumn{1}{c}{P} & \multicolumn{1}{c}{R} & \multicolumn{1}{c}{F1} & \multicolumn{1}{c}{P} & \multicolumn{1}{c}{R} & \multicolumn{1}{c}{F1} & \multicolumn{1}{c}{Acc} & \multicolumn{1}{c}{P} & \multicolumn{1}{c}{R} & \multicolumn{1}{c}{F1} \\ \midrule
MNB & 0.902 & 0.873 & 0.888 & 0.854 & 0.935 & 0.892 & 0.379 & 0.027 & 0.050 & 0.874 & 0.859 & 0.874 & 0.860 \\
RF & 0.875 & 0.881 & 0.878 & 0.862 & 0.916 & 0.888 & 0.333 & 0.005 & 0.010 & 0.866 & 0.844 & 0.866 & 0.851 \\
SVM & 0.926 & 0.844 & 0.883 & 0.881 & 0.914 & 0.897 & 0.211 & 0.330 & 0.257 & 0.861 & 0.877 & 0.861 & 0.868 \\
fastText & 0.947 & 0.890 & 0.917 & 0.905 & 0.955 & 0.929 & 0.463 & 0.177 & 0.256 & 0.891 & 0.883 & 0.891 & 0.883 \\
DistilBERT & 0.932 & 0.918 & 0.925 & 0.913 & 0.934 & 0.923 & 0.364 & 0.312 & 0.336 & 0.904 & 0.901 & 0.904 & 0.902 \\
BERT & 0.933 & 0.927 & 0.930 & 0.913 & 0.940 & 0.926 & 0.387 & 0.261 & 0.312 & 0.909 & 0.903 & 0.909 & 0.905 \\
RoBERTa & 0.933 & 0.931 & 0.932 & 0.919 & 0.941 & 0.930 & 0.386 & 0.269 & 0.317 & 0.912 & 0.906 & 0.912 & 0.909 \\
XML-RoBERTa & 0.929 & 0.930 & 0.930 & 0.922 & 0.935 & 0.928 & 0.359 & 0.288 & 0.320 & 0.909 & 0.905 & 0.909 & 0.907 \\ \bottomrule
\end{tabular}
\caption{Experimental results on \textbf{Task 3: Ternary classification (PNN).} 
% textcolor{red}{some mismatch in combining the tables-to be checked yet}.
}
\label{tab:result_task3}
\end{table*}
%%%%%%%%%%%%%%

\subsection{Task 3: Ternary classification (PNN)}
Table \ref{tab:result_task3} provides the experimental results on task 3, where the models have to differentiate among \textit{positive}, \textit{negative}, and \textit{neutral} reviews. Similar to previous two tasks, transformers produced better results compared to the classical and deep learning based methods. 

As can be seen in the table, better results are reported for all the methods on \textit{positive} and \textit{negative} classes. However, the performance of the proposed methods is significantly lower especially for the bag of words and ngram with the Naive Bayes classifier. One of the main reasons for the lower performance on \textit{neutral} class is due to the fewer samples in the class as described in Section \ref{sec:dataset}. We note that task 2 and task 3 are performed separately to analyze the impact of the fewer samples in the \textit{neutral} class.

\section{Discussion}
\label{sec:discussion}
Contact tracing of COVID-19 patients has been globally recognized as one of the most effective ways of controlling the infection rate. However, there are several limitations of the existing mechanisms. Manual contact tracing is a tedious and time-consuming process. Moreover, it is difficult to keep track of all potential contacts of a patient. Digital solutions, such as the use of mobile applications, has been considered as a promising solution where a patient's contacts can be traced and informed quickly. However, there are several concerns over the working mechanism and performance of the applications. This work has revealed different facets of the COVID-19 contact tracing applications, advantages, drawbacks, and users' concerns over these applications. We have summarized the main points hereafter. 

%\jq{More generally, is the focus of our paper on sentiment analysis for contact tracing in general or for particular contact tracing app. If the latter, then the sentiment is going to be influenced by technical factors of course. If the former, we should compare our insights against those from previous works and note how we agree and differ from the existing theoretical and empirical works.}

\begin{itemize}
    \item  The idea/initiative of contact tracing via a mobile application is highly appreciated by people worldwide. Besides contact tracing, the applications are also proved useful in implementing and ensuring public policies on COVID-19. However, there are also some concerns over the working mechanism and the effectiveness of the applications. 
    \item Analysis of users’ reviews on these applications helps to better understand and rectify the concerns over the applications.  
    \item Majority of the reviews lie in three categories, namely \textit{positive}, \textit{negative}, and \textit{technical issue}. On the other hand, very few neutral reviews are observed.
    \item Privacy in terms of tracking via GPS and access to the gallery and other information by the applications have been the main concerns. Moreover, a vast majority of the users of these applications in different parts of the world are not with the high power consumption of the applications.
    \item Majority of the users also faced some technical problems while using the applications. Some key technical issues include device compatibility, registration, slow updates, connectivity issues, and lack of support of some languages e.g., English. 
    \item The distribution of \textit{negative}, \textit{positive}, \textit{neutral}, and \textit{technical issues} may vary over time. 
    \item Overall better performance has been observed for the AI models in sentiment analysis of users' reviews allowing an efficient analysis of users' response to the application more quickly. 
    \item The transformers have been proved more effective among the models deployed for sentiment analysis in this work. 
    \item The experimental results indicate that reviews highlighting technical problems in the applications evoke negative emotions/sentiments.
\end{itemize}
\section{Conclusions}
\label{sec:conclusions}
In this paper, we focused on the sentiment analysis of users' reviews on the COVID-19 contact tracing mobile applications and analyzed how users react to these applications. To this aim, a pipeline is composed of multiple phases, such as data collection, annotation via a crowd-sourcing activity, and development, training, and evaluation of AI models for the sentiment analysis. The existing literature mostly relies on the manual/exploratory analysis of users' reviews on the application, which is a tedious and time-consuming process. Moreover, in the existing studies, generally, data from fewer applications are analyzed.  In this work, we showed how the automatic sentiment analysis can help in analyzing users' responses to the application more quickly. Moreover, we also provided a large-scale benchmark dataset composed of 34,534 reviews from 47 different applications. We believe the presented analysis and the dataset will support future research on the topic. 

We believe, many interesting applications and analysis can be conducted keeping the dataset as a baseline. Temporal and topical analysis are the key aspects to be analyzed in the future.

\bibliographystyle{unsrt}
\bibliography{References}

\end{document}